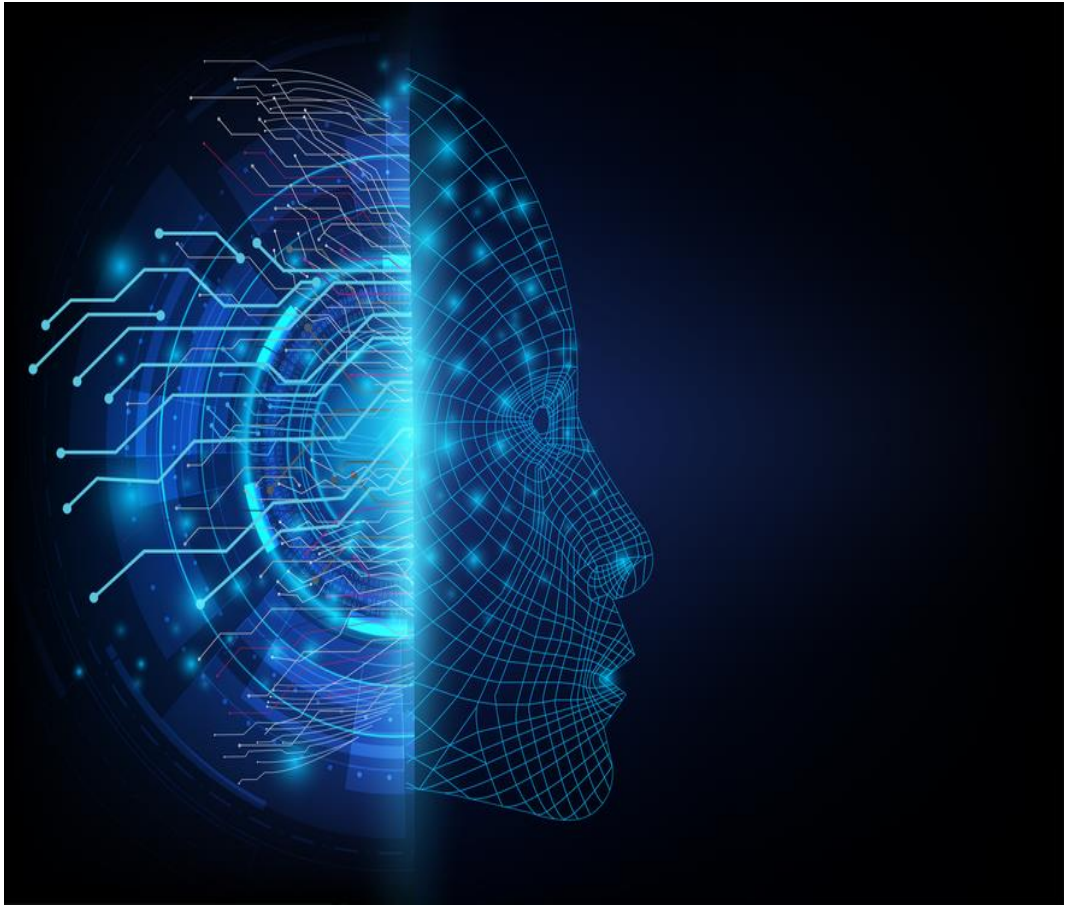

# AUTO-ENCODING A KNOWLEDGE GRAPH USING A DEEP BELIEF NETWORK

A Random Fields Perspective

ABSTRACT
We started with a knowledge graph of connected entities and descriptive properties of those entities, from which, a hierarchical representation of the knowledge graph is derived. Using a graphical, energy-based neural network, we are able to show that the structure of the hierarchy can be internally captured by the neural network, which allows for efficient output of the underlying equilibrium distribution from which the data are drawn.

Robert A. Murphy
robert.a.murphy@wustl.edu

# Table of Contents

# Introduction

There are times when there may be a need for expansion of the knowledge obtained from a collection of documents that may have been scanned/stored using optical character recognition (OCR) or some other methodology. Given that insights from the corpus of documents may be sparse, we might want to grow the data set and one way to do that is domain adaptation via transfer learning, which amounts to using a trained model from another domain and applying that model to the domain of interest.

Using a large, pretrained, global vectors for word representations (GloVe) data set, we have a general embedding of the sparse data set into a chosen "universe" of knowledge and by using a deep belief network (DBN), we will obtain a specific embedding into the space containing the sparse insights obtained from the corpus of documents. Then, other services are applied on the large data set to obtain a knowledge graph (KG) from the expanded data set such that hierarchical clustering can be used for searching entities (feature values) in the KG, as we might want to trace child entities through properties of a parent entity.

# GloVe for Knowledge Expansion

We assume that a corpus of documents has been obtained and the processes of tokenization (word/phrase separation), stemming (equating participles like run, ran, running), entity extraction (identification of people, places, events), and part-of-speech (noun, verb, adjective, adverb, etc.) identification has already been done. Learning the distribution of word-word co-occurrences requires us to first form something similar to a document term matrix (DTM) for a corpus, wherein a list of documents are the rows of the matrix and all words that can appear in a given document are the columns. The values in a given row/column of the matrix are the number of times that the given word appears in the corresponding document.

Likewise, in a matrix of word-word co-occurrences, the rows are also the listing of words that appear in all the documents, except the values in a given row/column indicates the number of times that the words appear in a document together. Considering a given row/column of the word-word co-occurrence matrix (i.e. a fixed word) and allowing the other words in each column of the row to vary over all possible values, amounts to the consideration of a conditional distribution in 2 variables with one being fixed. As such, if we learn the conditional distribution associated with the given row/column, then we use the typical set up where we consider the linear combination of the values in the row/column as inputs to an exponential function to capture non-linearities in the distribution.

Given that most values in the word-word co-occurrence matrix will be zero, we can simplify the model of the data by assuming that the joint distribution is the product of its marginals, termed **separability**. This assumption is safe, given that the matrix is sparse, leading to an asymptotically finite mean/variance of each marginal. Now, the joint distribution is easily seen to be log-multilinear, since the log exposes the energy model using the inputs, which is linear in all terms, when all but one term is fixed.

For each row of the matrix, if we seek its marginal distribution, then we are effectively trying to find the set of words (out of all possible words, as indicated by the corpus) that are likely to appear, amounting to the so-called **continuous-bag-of-words** GloVe problem. Performing the same set up while seeking the marginal distributions of the columns amounts to the **skip-gram** GloVe problem, whereby we are given the surrounding words and we want to predict the missing word that has been skipped.

Given the sparse nature of our original matrix, we could make use of its singular value decomposition (SVD) to find orthogonal matrices that pre and post multiply a (square) reduced-rank matrix to produce our sparse matrix. The reduced rank matrix is our embedding of words into a lower dimensional subspace and the rows/columns define the "universe" of the subspace, with the corresponding words that define the rows/columns being our effective knowledge base. Note that the generic embedding gives us a many-to-one relation, whereby a vector containing many values is reduced to a vector containing fewer values, e.g. $(q, r, s, t, u, v, w, x, y, z) \mapsto (r, t, u, v, x)$. The resulting knowledge base should contain the space that contains our sparse knowledge base, allowing the sparse base to grow within it.

Since we already have our generic embedding, all that is left to define is the sparse embedding, which is just a word-word correspondence between the set of words from the sparse data set into the larger space. This is a simple matter of equating words from the sparse data set to a subset of the generic (larger) data set so that when the knowledge graph is produced using the generic data set, the vertices, edges and properties of the sparse data set are known in terms of the generic data set. Finally, note that one of the edge properties of the knowledge graph has a natural value that's determined by the co-occurrence relationships with other words in the corpus and the number of times both occur together, which is defined in the original matrix.

## The Approach

Recall that a **knowledge graph** is an undirected, graphical representation of data relationships that can be further represented as a mixed, binary hierarchy, where each branch in the hierarchy flows from some property of the data. As such, the first level consists of all the data, while in the second level there are sibling branches, each given by a particular property. Successive levels following the 2[nd] level consist of binary branches on the given property, followed by sibling branches on all other properties, followed by binary branches on the remaining individual properties, ..., until all properties are exhausted along each path.

The entire hierarchy is encoded using a deep belief network (DBN) consisting of a determinative number of restricted Boltzmann machine (RBM) layers. For each layer containing sibling branches given by each of the M properties, its corresponding RBM consists of $2^{(J-2)/2} * M^{J/2}$ linear combiners in a hidden layer utilizing the Gibbs distribution as its activation function, with an identity function at its output layer, where J is the layer of the DBN. For each layer consisting of binary child branches that follow sibling levels, its corresponding RBM consists of $2^{(J-1)/2} * M^{(J-1)/2}$ linear combiners in a hidden layer utilizing the Gibbs distribution as its activation function, with an identity function at its output layer, where J is the layer of the DBN.

Each neuron in each of the binary levels takes as inputs, all the data from exactly one of the leaf nodes in its preceding sibling level. Each is a linear combiner utilizing the Gibbs distribution as its activation function, with an identity function at the output layer. Note that the number of neurons in this level has the same effect as defining a hyper parameter η that can be chosen by the user (or adjusted on the fly) to achieve a certain balance of class members within each leaf node in the binary level that follows a sibling level. Then, each leaf node in a sibling/parent level is split into binary classes by $\{x : x < \eta\}$ or

$\{x : x \geq \eta\}$ in the binary/child level. The levels stop when all branch properties are exhausted in each path.

Now for each new data point, the functions learned by the deep learner at each level will determine the path taken. As a result, each of the leaf nodes hosting the data point along the branch paths are also determined.

## Deep Belief Network

### Random Field Setup

Imagine a closed, bounded region in the 2-dimensional plane, partitioned by uniformly spaced, orthogonal, vertical and horizontal lines. At the intersections of the orthogonal lines, a process P independently generates points according to some unknown probability distribution. Two points connect to each other (and the edge between them is open) with some probability p whenever they occupy neighboring intersections in the partition. Define a **clique** to be a cluster of intersections in a subset of the partition and define a **conditional specification** to be the set of conditional probability distributions over edge values in all cliques in the partition, when all other edge values outside each clique are fixed.

Allowing for open edges to have a value of 1 and closed edges to have a value of -1, a clique has an associated energy H that can be computed as a function of its edge values. As such, each distribution in the conditional specification can be expressed as a function of its energy. A **deep belief network** is a multi-hidden layer exploitation of the connected graph structure (and associated energy-based conditional specification). It seeks to employ stochastic gradient descent (optimization along each noisy coordinate) to find an estimate of the global distribution (**random field**) across all intersections in the partition that locally approximates the elements within the conditional specification, the very definition of a greedy algorithm.

All that is left to handle is a multi-dimensional data set, since we started with the assumption of a bounded region within a 2-dimensional plane, when hosting the data points generated by the random process, P. However, in Murphy [Neural Network Support Vector Detection via a Soft-Label, K-Means Classifier, arXiv, 2016], we can define a uniform partition of a bounded, unit area region using the number of data points and its presumed number of connected clusters. Then, we are guaranteed the existence of a canonical projection of the higher dimensional data points into the 2-dimensional plane that preserves cluster membership. Therefore, without loss of generality, we may assume that the original data points and bounded region are within the 2-dimensional plane.

## Random Field Illustration

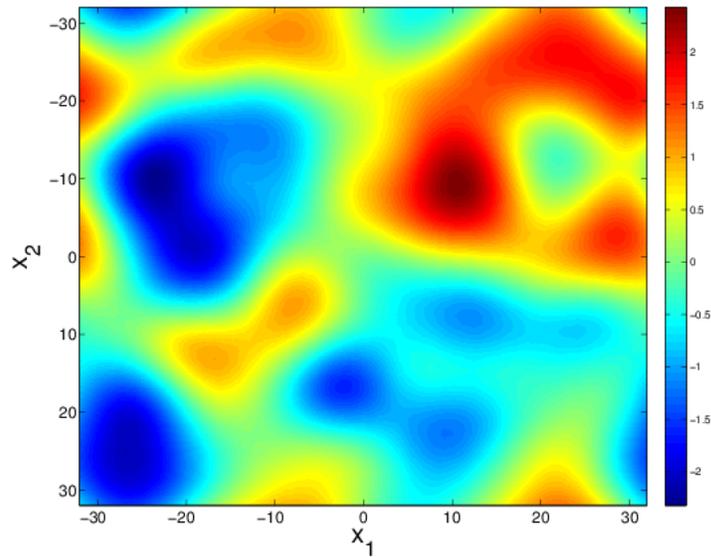

In the preceding image, the bounded 2-dimensional region is partitioned with a resolution matching that of the pixels.  From the preceding section, we know that the red and blue clusters are representative of clusters of higher dimensional data points that have been maintained after canonical projection.  In blue clusters, lighter pixels indicate weaker correlation to their darker blue counterparts, but stronger correlation with those nearer to them.  Similarly, in red clusters, yellow and orange pixels are less correlated with brighter red pixels, but more strongly correlated with each other.

Relative strength of connection is indicated by edge probability, p, which underscores the likelihood that two representative pixels are correlated.  Since points are independently and identically generated by the process P, it is easy to see that points on the outer fringes of a cluster are less correlated with ones nearer the center, by simply applying the product probability measure on paths of edges between pixels near the center of a cluster to pixels near the fringes of a cluster.  Likewise, shorter paths between pixels both near the center or both near the fringes indicates a stronger correlation using the product rule.  Hence, their colors are closer in hue.  Intermediate green voids are unpopulated regions of the point process, P.

Connected clusters of pixels are within regions of cliques.  The color coding follows some probability distribution that matches the distribution of the originally-correlated, higher dimensional data points.  Given the color coding of pixels that are external to a given clique, an element of the **conditional specification** is the distribution over edge values within the clique.  Then, our **random field** associated to the original data points is the distribution across the entire partition, which locally produces the color coding that has been observed in the image.

## Restricted Boltzmann Machine

In each layer of the network, multiple local approximations detail a belief about the nature of the global distribution and each layer informs its successor.  As each individual layer in the network models local connectivity of an undirected graph, necessarily being indicative of feedback, we may consider each of the layers as a self-contained network, with input from its predecessor, followed by a hidden layer (providing feedback) and output that informs its successor network.

Each of the elements of the conditional specification is itself a Gibbs distribution as a function of an energy model so that each neuron in the hidden layer makes an approximation of the global distribution.  Since a single layer in the overall network only maintains connections between its visible and hidden neurons, making each layer a **restricted Boltzmann machine**, then, by definition, a **deep belief network** is a multi-hidden layered network of individual RBMs.

## The Encoding

Since the DBN consists of layers of RBMs, then the energy function that defines the random field over all edges in the partition is known, up to a set of modeling constants, as the one-term (multilinear) products of individual coordinates from inputs at the visible and hidden layers in each RBM.  Each RBM in the DBN is akin to a multi-layered perceptron with a single hidden layer and Gibbs distribution as its activation function.  As such, the DBN can be coded with dense layers of artificial neural networks (ANN) utilizing this special activation function, along with mean squared error as the loss function and stochastic gradient descent as the optimizer.

### Deep Learning Framework Pseudo Code

Model the deep learner as a linear stack of sequential layers;

Define M to be the dimension of the input data;

Add a dense layer with tensor shape accepting M columns and any number of rows … a rectified linear unit activation function … with M outputs;

Initialize odim = 2;
For J in the range of values from 1 to M :
    Define input dim = 2 * odim;
    Add a dense layer with tensor shape accepting input odim columns and any number of rows … a scaled exponential linear unit (Gibbs distribution) activation function … with dim outputs;
    Define output odim = 2 * dim;
    Add a dense layer with tensor shape accepting input dim columns and any number of rows … a scaled exponential linear unit (Gibbs distribution) activation function … with odim outputs;

Add a dense layer with tensor shape accepting odim columns and any number of rows … scaled exponential linear unit (Gibbs distribution) activation function … with M outputs;

Compile the model with a mean squared error loss function and a stochastic gradient descent optimizer;

Fit the inputs to the inputs with a predetermined number of epochs of training data iterations, as this is an autoencoding of the original inputs;

### Keras Example

*Specifications*

The algorithm in the preceding section was coded using Python version 2.7.12 and Keras version 2.2.4 with Tensorflow backend version 1.12.0 running on 64-bit Linux mint version 18.2 utilizing an Intel i7-8650 multi-core processor, Nvidia GeForce GTX 1060 graphics processor and 16GB RAM.

*Network Structural Diagram*

For this example, suppose the number of properties is M = 2.  From section "Approach", we will have 2 sibling layers and 2 binary layers that flow from/to the sibling layers, in an alternating fashion such that the structure of the network consists of 2 distinct RBMs comprised of layers 1-3 and layers 3-5.  The final output layer takes inputs from the output of the final RBM so that the structure of the DBN consists of 2 RBMs followed by an output layer.  From the same section, we know that the number of neurons in each of the sibling layers is $2^0*(2)$, $2*(2)^2$.  Likewise, in the binary layers, the number of neurons is $2*(2)$, $2^2*(2)^2$.

The network in this example will have the following structure:

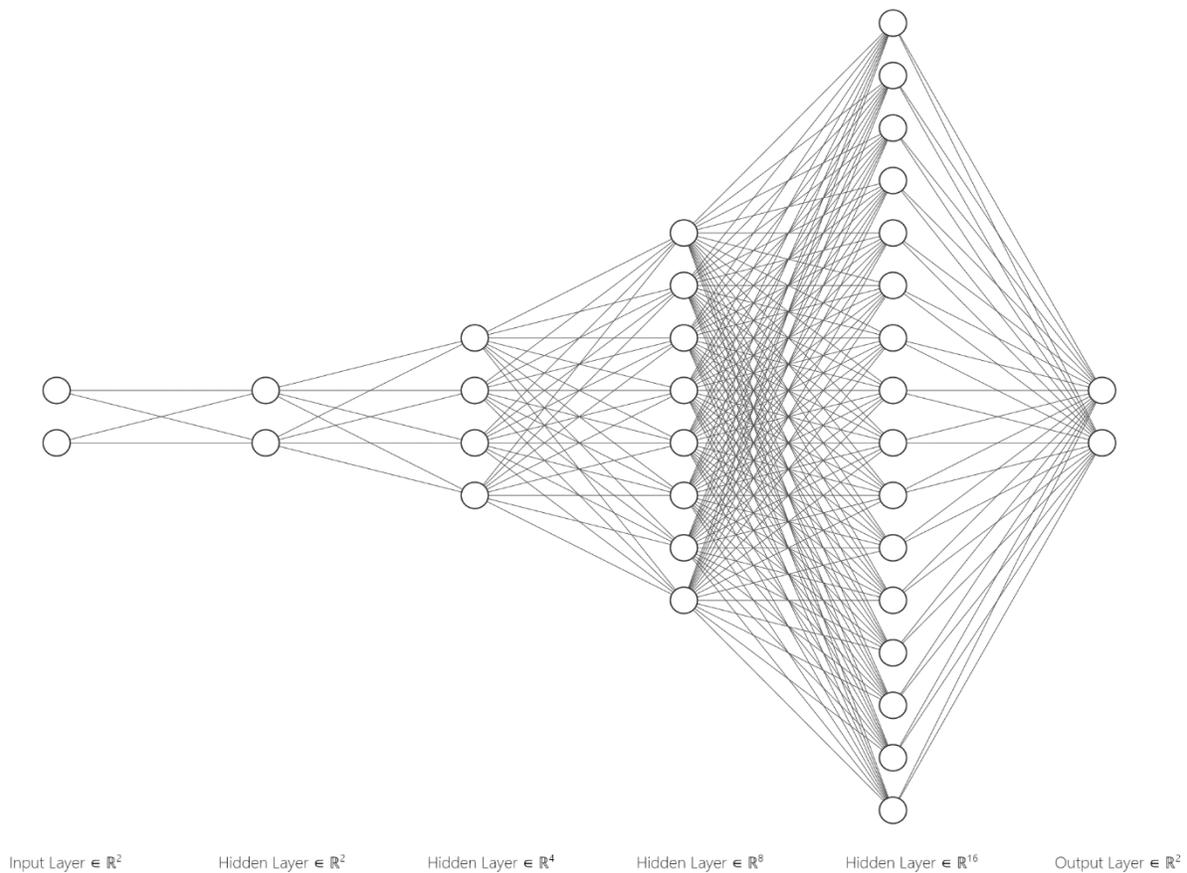

Input Layer $\in \mathbb{R}^2$    Hidden Layer $\in \mathbb{R}^2$    Hidden Layer $\in \mathbb{R}^4$    Hidden Layer $\in \mathbb{R}^8$    Hidden Layer $\in \mathbb{R}^{16}$    Output Layer $\in \mathbb{R}^2$

Properties of the knowledge graph are a continuum of values, as opposed to categorical.  If the data in each of the binary levels are approximately uniformly distributed over the range of values of a given property, then it is easy to see that neurons become more specialized to a portion of the property values as the data progresses through the hidden layers of the network. In general, as the number of properties increases, more neurons are required to accomplish the encoding and each neuron in the network must become more specialized to a portion of the knowledge graph. Finally, note that at the input level, the number of neurons matches the number of neurons in the first hidden level, as the inputs are values of properties. As each property figures prominently in the network construction that follows, feature selection is not allowed to occur, as all properties contain specific content that is useful when building the model. Thus, herein lies an instance where the input layer is not allowed to behave like a Gabor filter (feature selector).

In summation, the neural network takes input and does not perform feature selection. Internally, the structure of the network captures the hierarchical representation of the knowledge graph, then outputs an equilibrium distribution that models the underlying subspace of the original, noisy data set.

*Results*

Setup

We used 100 epochs for training on 500 rows of 2-dimensional data, drawn uniformly from the interval, [0,1]. As the properties are continuous, we only want to test if the autoencoder generates data from the underlying distribution from which the noisy input data are drawn. As such, the strategy employed is to generate uniformly distributed training and test data sets, then employ a statistical test to gauge whether the output data is drawn from the equilibrium (limiting) distribution of the test data. For testing, 50 rows of 2-dimensional data were independently drawn, uniformly from the interval, [0,1].

Shapiro-Wilke Test

Recall that the exponential distribution is the limiting distribution of a sequence of uniformly distributed random variables and the output of the DBN is a model of an exponential distribution. The **Shapiro-Wilke Test of Normality** is a statistical test that gauges how much a data set deviates from an equilibrium state when a perturbation (ordering) is applied to the original data set. The test was originally developed for determining if a given data set was drawn from a normal distribution, since the normal distribution is the limiting, equilibrium distribution for any sequence of data having asymptotically finite mean and variance. However, the test can be applied to any distribution from the exponential family. Thus, the Shapiro-Wilke test essentially answers the question of how close to reality is an assumption of a steady, unchanging state (stationarity), when applied to the given data set.

The Shapiro-Wilke test outputs a W-metric that is the ratio of the amount of variability in the ordered data over the amount of variability in the originally-unordered data. The closer the metric is to 1 is an indication that the original data set was more likely drawn from an equilibrium distribution, as applying an order statistic does not introduce significantly more variability.

The test data indicates W = 0.93503, while the model output, using the test data, indicates a value of W = 0.98114. We can reasonably infer that the model is finding the underlying equilibrium distribution from which the noisy test data are drawn.

# Closing

We started with a knowledge graph of connected entities and descriptive properties of those entities, from which, a hierarchical representation of the knowledge graph is derived. Using a graphical, energy-based neural network, we are able to show that the structure of the hierarchy can be internally captured by the neural network, which allows for efficient output of the underlying equilibrium distribution from which the data are drawn.

# Bibliography

[1] Crempien, Jorge & Auricchio, Ferdinando & Lai, Carlo & Nobile, Fabio. (2019). *Response Statistics of Uncertain Dynamical Systems Subjected to Stochastic Loading Using Sparse Grid Collocation Techniques*, Figure 2.1.

[2] Dialani, Priya (2019). *Artificial Intelligence is a Great Detector Tool*, Artificial Intelligence Latest News, https://www.analyticsinsight.net/artificial-intelligence-is-a-great-detector-tool/, Page 1 Main Figure.

[3] Grimmett, Geoffrey (1999), *Percolation*, Springer-Verlag.

[4] Grimmett, Geoffrey (2006), *The Random-Cluster Model*, Springer-Verlag.

[5] Guyon, Xavier. (1995), *Random Fields on a Network: Modeling,Statistics and Applications*, Springer.

[6] Pennington, Jeffrey & Socher, Richard & Manning, Christopher D. (2019). *GloVe: Global Vectors for Word Representation*, https://nlp.stanford.edu/projects/glove/.

[7] Shapiro, S.S., Wilk, M.B. (1965), *An Analysis of Variance Test for Normality (Complete Samples)*, Biometrika, Volume 52, pp. 591 - 611, 1965.